\documentclass[runningheads]{llncs}

\usepackage[T1]{fontenc}
\usepackage{graphicx,verbatim}
\usepackage{booktabs}
\usepackage{amsmath}
\usepackage{amssymb}
\usepackage{comment}
\usepackage{marvosym}
\usepackage[hidelinks]{hyperref}
\begin{document}

\title{Federated LoRA Adaptation of BiomedCLIP Across Four International Chest X-Ray Cohorts}
\titlerunning{Federated Adaptation of Biomedical VLMs}

% \title{Federated LoRA Adaptation of BiomedCLIP Across
% Four International Chest X-Ray Cohorts}

% \titlerunning{Federated Adaptation of Biomedical VLMs}

\authorrunning{Poudel et al.}
\author{Sanjaya Poudel\textsuperscript{1} \and
Nirajan Kunwor\textsuperscript{2} \and
Manish Dhakal\textsuperscript{3} \and
Debesh Jha\textsuperscript{4} \and
Sunil Kumar Gaire\textsuperscript{1\,(\Letter)}}
\institute{\textsuperscript{1}North Carolina A\&T State University, \textsuperscript{2}Tribhuvan University, \textsuperscript{3}University of Tennessee-Knoxville, \textsuperscript{4}University of South Dakota 
\\
{\small (\Letter)~\email{skgaire@ncat.edu}}
}

\maketitle

%% Note: This version is shared with Prof. Gaire for Review

\begin{abstract}
  Federated learning (FL) lets institutions train a shared model without exchanging data, and Low-Rank Adaptation (LoRA) makes this practical at scale by communicating only compact low-rank updates. Biomedical imaging is a compelling setting for this combination: patient data are archived behind privacy regulations, and institutions differ widely in scanners, protocols, and compute. Such heterogeneity raises the question of how federated LoRA updates should be aggregated, increasingly pressing as multimodal vision-language models become central to medical image analysis. We benchmark federated Parameter-efficient fine-tuning (PEFT) of BiomedCLIP for chest radiograph classification across four public cohorts on three continents (USA, Vietnam, Spain). Federated LoRA adaptation improves shared-class AUC on all four cohorts over the unadapted BiomedCLIP backbone (mean $0.687 \rightarrow 0.802$), showing that the gains come from federated adaptation rather than from the pretrained model's zero-shot ability. Relative to isolated single-cohort training, federation improves the weaker cohorts while largely preserving the strongest and approaches a centralized reference (0.812) that pools all data. The singular value decomposition (SVD)-based product-space aggregation introduced by FlexLoRA is essential to this gain (naive factor averaging drops mean AUC by 0.097), whereas a drift-correcting optimizer (FedProx) shows no benefit over FedAvg in our single-seed runs, consistent with LoRA's low-rank updates already limiting client drift. Biomedical vision-language models can thus be adapted collaboratively across heterogeneous, geographically distributed institutions without centralizing data.Code is available at: \href{https://github.com/GaireLaboratory/FedLoRA-BiomedCLIP}{github.com/GaireLaboratory/FedLoRA-BiomedCLIP}
  \keywords{Federated Learning \and LoRA \and Vision-Language Models.}
\end{abstract}

\section{Introduction}
\label{sec:intro}
Medical imaging data are fragmented across institutions by privacy regulations, so a model trained at any single site observes only a narrow slice of the patient population. Federated learning (FL) offers a way forward, training a shared model while raw images stay local and only model updates are exchanged~\cite{ref_fedavg}. This avenue is especially compelling for biomedical foundation models: multimodal vision-language models (VLMs) such as BiomedCLIP~\cite{ref_biomedclip} attain strong performance by aligning images with text, yet adapting them to clinical tasks benefits from large, diverse data that no single institution holds. Federating their adaptation could combine cohorts across sites without moving data, making FL a natural setting for multimodal biomedical analysis.

Fully fine-tuning such models in federation is impractical due to communication
cost and overfitting on small institutional datasets. Parameter-efficient
fine-tuning (PEFT), in particular Low-Rank Adaptation (LoRA)~\cite{ref_lora},
instead updates only small low-rank matrices, cutting per-round communication by
over $99\%$ and making it well suited to FL. Each client's LoRA update, however, is stored as two small matrices whose
\emph{product} forms the actual weight update. Averaging these matrices
separately across clients does not give the average of their products, so the
aggregated update no longer represents what the clients actually learned. FlexLoRA~\cite{ref_flexlora} addresses this by
reconstructing each client's full-size update, averaging the reconstructed
weights, and redistributing low-rank factors through singular value decomposition
(SVD). We adopt this SVD-based aggregation but hold the rank fixed across clients,
isolating its aggregation-correctness benefit; rank-heterogeneity handling is a
direct avenue for future scaling.

These federated-LoRA schemes, however, have been validated almost exclusively on
natural-language tasks~\cite{ref_flexlora,ref_fedsa,ref_fedex}; their behavior on
biomedical VLMs where the objective is image-text contrastive alignment and
clients differ by protocol, geography, and label vocabulary remains unexamined.
Prior federated chest radiograph studies use convolutional classifiers with fixed
label heads~\cite{ref_tayebi,ref_surgical}, motivating a systematic benchmark of
federated PEFT for medical VLMs under SVD-based aggregation.

We present a cross-continental study of federated PEFT of BiomedCLIP for chest
radiograph classification, federating LoRA adapters with FlexLoRA's
SVD-based aggregation across four public datasets on three continents: NIH
ChestX-ray14 and CheXpert (USA), VinDr-CXR (Vietnam), and PadChest  (Spain). 
Our
contributions are:
\begin{itemize}
  \item We provide, to our knowledge, the first systematic benchmark of federated
    LoRA-based PEFT for a biomedical VLM across four real, geographically distinct
    chest X-ray cohorts; the benchmark itself, rather than a new algorithm, is our
    primary contribution.

  \item Federated LoRA adaptation improves shared-class AUC over the frozen BiomedCLIP backbone on all four cohorts (mean $0.687 \rightarrow 0.802$), confirming the adaptation not the pretrained backbone alone drives the gains.

  \item Relative to each cohort's strongest single-client baseline, federation
   improves the weaker cohorts (CheXpert $+0.038$, VinDr $+0.016$), ties NIH,
    and slightly reduces the strongest (PadChest), raising mean shared-5 test
    area under the curve (AUC) from $0.776$ to $0.802$ (FedAvg) and approaching
    a centralized reference ($0.812$) without pooling data.
  \item We show SVD-based aggregation is essential in one-shot merging:
    naive factor averaging drops mean shared-5 AUC by $0.097$, to the level of
    the frozen backbone.
  \item We compare FedAvg and FedProx under LoRA and observe no meaningful
    benefit from the proximal term ($0.802$ vs.\ $0.799$), consistent with LoRA
    already limiting client drift.
\end{itemize}

\section{Related Work}

\paragraph{Federated Learning in Medical Imaging:}
FL enables collaborative training without sharing patient data. FedAvg~\cite{ref_fedavg},
FedProx~\cite{ref_fedprox}, SCAFFOLD~\cite{ref_scaffold}, and FedBN~\cite{ref_fedbn}
address optimization under heterogeneous data~\cite{ref_fedsurvey}. FL has been applied to chest
radiograph analysis, including with differential
privacy~\cite{ref_tayebi,ref_surgical,ref_fedmed,ref_dp}, but most studies use
convolutional networks rather than multimodal foundation models.

\paragraph{Parameter-Efficient Fine-Tuning:}
PEFT updates only a small subset of parameters. LoRA~\cite{ref_lora} is the most
widely adopted, with demonstrated effectiveness in medical
imaging~\cite{ref_poudel2026}. FedIT~\cite{ref_fedit} first combined LoRA with
federated averaging, and later work refined aggregation of low-rank updates under
heterogeneous clients~\cite{ref_flexlora,ref_fedsa,ref_fedex,ref_ffalora}.

\paragraph{Vision-Language Models in Medical Imaging:}
VLMs learn aligned image-text representations through contrastive pretraining. Following CLIP~\cite{ref_clip}, medical variants such as MedCLIP~\cite{ref_medclip}, CheXzero~\cite{ref_chexzero}, BioViL~\cite{ref_biovil}, and BiomedCLIP~\cite{ref_biomedclip} perform strongly. Federated VLMs have been explored with adapters and foundation models~\cite{ref_fedclip,ref_fedfms,ref_vlsmadapter}, but federated PEFT of biomedical VLMs remains largely unexplored the gap we address.
%% ============================================================
%% METHODS  (owners: Nirajan, Sanjay)
%% ============================================================
\section{Methods}
\label{sec:methods}

\subsection{Overview}

\begin{figure}[h]
  \centering
  \includegraphics[width=0.60\textwidth]{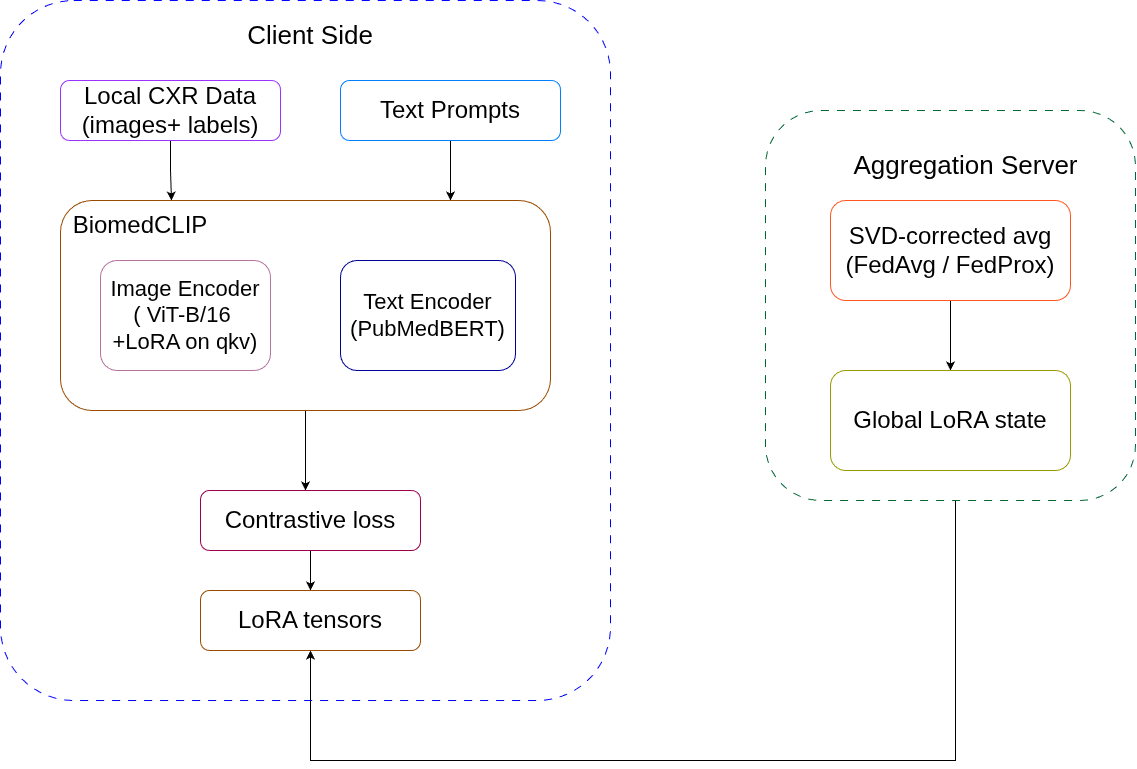}
  \caption{Overview of the federated pipeline.}
  \label{fig:architecture}
\end{figure}

We study federated PEFT of a biomedical vision-language model across four chest
radiograph cohorts through three experiments of increasing collaboration:
(i) \emph{single-client} baselines, where each cohort trains in isolation;
(ii) \emph{one-shot aggregation}, where the four locally trained adapters are
merged once; and (iii) \emph{multi-round federation}, where clients iteratively
train and aggregate over five rounds. Raw images never leave their source cohort;
only adapter weights are exchanged (Fig.~\ref{fig:architecture}).

\subsection{Model Architecture and LoRA Fine-Tuning}

We use BiomedCLIP (frozen PubMedBERT text encoder, ViT-B/16 image encoder
pretrained on 15M biomedical figure-caption pairs). LoRA adapters~\cite{ref_lora} on the fused query-key-value (\texttt{qkv})
projection of each transformer block (rank $r=8$, scaling factor $\alpha=16$, dropout $0.1$) yield
${\sim}0.25\%$ trainable parameters, reducing the
per-round payload to ${\sim}1.13$\,MB per client (>99\% smaller than full-model
exchange).
Each client locally optimizes its LoRA adapters using an image--text contrastive loss:
\begin{equation}
  \mathcal{L}_{\text{CLIP}} = \tfrac{1}{2}\left[\mathcal{L}_{i\to t}(\tau^{-1}\mathbf{Z}_i \mathbf{Z}_t^\top) + \mathcal{L}_{t\to i}(\tau^{-1}\mathbf{Z}_t \mathbf{Z}_i^\top)\right]
\end{equation}
where $\mathbf{Z}_i$, $\mathbf{Z}_t$ are normalized image and text embeddings,
$\tau=0.07$ is the temperature, a fixed scaling constant standard in contrastive
vision-language training~\cite{ref_clip}, which sharpens the softmax over cosine
similarities (equivalent to a logit scale of $1/\tau \approx 14.3$), and $\mathcal{L}_{i \to t}$,
$\mathcal{L}_{t \to i}$ are symmetric cross-entropy losses. Prompts list up to three of each image's present findings:
``Chest radiograph showing $\{$finding$_1$, finding$_2$, finding$_3\}$'', or
``Chest radiograph showing no findings'' for negatives.

\subsection{Federated Aggregation with SVD-based LoRA}

After each communication round, only LoRA tensors are exchanged. Each client $k$
learns a pair of low-rank LoRA matrices a down-projection
$A_k \in \mathbb{R}^{r \times d}$ and an up-projection
$B_k \in \mathbb{R}^{3d \times r}$ with rank $r \ll d$ whose scaled product
$\Delta W_k = (\alpha/r)\,B_k A_k$ forms the update to the fused \texttt{qkv}
target weight $W \in \mathbb{R}^{3d \times d}$ (the projection maps the
$d$-dimensional token to stacked query, key, and value). Since averaging factors
independently is not equivalent to averaging products
($\bigl(\tfrac{1}{N}\sum_k B_k\bigr)\bigl(\tfrac{1}{N}\sum_k A_k\bigr) \neq
\tfrac{1}{N}\sum_k B_k A_k$), we follow FlexLoRA~\cite{ref_flexlora} and aggregate
in product space: we average the reconstructed updates $B_k A_k$ across clients and
project the result back to rank $r$ via truncated SVD. Because the average of $N$
rank-$r$ updates can have rank up to $Nr$, this rank-$r$ projection is an
approximation, not an exact reconstruction:
\begin{equation}
  \overline{\Delta W} = \sum_{k=1}^{N} w_k\, B_k A_k, \qquad U, S, V^\top = \mathrm{SVD}(\overline{\Delta W})
\end{equation}
\begin{equation}
  \overline{B} = U_{:,1:r}\,\mathrm{diag}(\sqrt{S_{1:r}}), \qquad \overline{A} = \mathrm{diag}(\sqrt{S_{1:r}})\,V^\top_{1:r,:}
\end{equation}

Here $\overline{\Delta W}\in\mathbb{R}^{3d\times d}$ is the weight-averaged update
($w_k=1/N$) and $U,S,V^\top$ its SVD; keeping the top $r$ components gives
$\overline{B}\,\overline{A}\approx\overline{\Delta W}$ at rank $r$. Although the rank-$r$ projection is lossy in principle, in practice it is nearly
information-preserving here: across the twelve adapted blocks the top-$r$
components retain on average $94.6\%$ of the averaged update's singular-value
energy (range $92.9$ - $95.4\%$), so the truncation discards only a small fraction
of the aggregated update.

\subsection{Federation Configurations}

We compare two algorithms with otherwise identical hyperparameters.
FedAvg~\cite{ref_fedavg} uses SVD-based LoRA aggregation with equal client weights
($w_k=1/N$) rather than the data-size weighting of the original formulation; our
one-shot ablation (Table~\ref{tab:aggregation}) shows equal weighting outperforms
size weighting (0.781 vs.\ 0.743), as size weighting lets the largest cohort
(CheXpert) dominate. Each client trains the contrastive loss locally. FedProx~\cite{ref_fedprox} adds a proximal
term $\tfrac{\mu}{2}\|\theta_k - \theta_{\text{global}}\|^2$ to the local
objective, where $\theta_k$ are client $k$'s current LoRA parameters,
$\theta_{\text{global}}$ the global LoRA parameters from the previous round, and
$\mu = 0.01$ the proximal coefficient regularizing each client toward the global
state. All experiments use seed 42 for reproducible splits and initialization.

%% ============================================================
%% EXPERIMENTS  (owners: Nirajan, Sanjay, Manish)
%% ============================================================
\section{Experiments}
\label{sec:experiments}

\subsection{Datasets and Preprocessing}

We use four public chest radiograph datasets (Table~\ref{tab:datasets}): NIH
ChestX-ray14~\cite{ref_nih}, CheXpert~\cite{ref_chexpert},
VinDr-CXR~\cite{ref_vindr}, and PadChest~\cite{ref_padchest}, from the USA,
Vietnam, and Spain. As native label vocabularies differ, we use the NIH and
CheXpert 14-label sets, the 14-class VinDr mapping, and the 30 most frequent
PadChest findings.

Images are resized to $224\times224$ and normalized using BiomedCLIP
preprocessing. PadChest 16-bit images are converted to 8-bit using 1st-99th
percentile normalization. CheXpert is restricted to frontal images and uncertain
labels ($-1$) are treated as positive (U-Ones~\cite{ref_chexpert}), applied
uniformly for simplicity though per-class policies can be preferable. PadChest is restricted to frontal adult
radiographs (PA, AP, AP\_horizontal; $\geq18$ years), consistent with prior
work~\cite{ref_tayebi}. We use each dataset's official test split where available;
for NIH we use the official patient-disjoint train/test lists, and for CheXpert
and PadChest we split by patient (80/10/10). As VinDr-CXR's official test labels
are not public, we split its official training set 80/10/10 at the image level,
since patient identifiers are unavailable.

\begin{table}[t]
\centering
\caption{Dataset partitions. Splits use seed 42. Patient-level splits are used where patient identifiers are available (CheXpert, PadChest).}
\label{tab:datasets}
\begin{tabular*}{\textwidth}{@{\extracolsep{\fill}}lrrrl@{}}
\toprule
Client (region) & Train & Val & Test & Split type \\
\midrule
NIH (USA)           & 77{,}872  & 8{,}652  & 25{,}596 & Patient-level (official) \\
CheXpert (USA)      & 153{,}182 & 18{,}958 & 18{,}729 & Patient-level \\
VinDr-CXR (Vietnam) & 12{,}000  & 1{,}500  & 1{,}500  & Image-level \\
PadChest (Spain)    & 65{,}173  & 8{,}078  & 8{,}147  & Patient-level \\
\bottomrule
\end{tabular*}
\end{table}

\subsection{Training and Evaluation Protocol}

Multi-round federation uses 5 communication rounds with 1 local epoch per round,
optimized with Adam ($\eta=2\times10^{-4}$, batch size 32). After each round,
client LoRA states are aggregated via SVD-based averaging and redistributed.
Single-client baselines train 5 epochs with early stopping on validation AUC,
matching federation's total local data passes.

We report macro-averaged AUC on each client's held-out test set in two modes:
full-class, over each dataset's native findings (14 for
NIH/CheXpert/VinDr, 30 for PadChest), and shared-5, over five findings
common to all four datasets (Atelectasis, Cardiomegaly, Consolidation, Effusion,
Pneumothorax), with each dataset's native label name mapped to the canonical
shared name.

This prompt construction is used only during training. At test time, each
shared finding is scored by the cosine similarity between the image embedding and
a single per-class text prompt (``Chest radiograph showing \{label\}''), and
macro-AUC is computed directly from these per-class similarity scores. Bootstrap 95\% confidence intervals (CIs) are computed from 300 test-set resamples and
reflect test-set, not training, variability. 

%% ============================================================
%% RESULTS
%% ============================================================
\section{Results}
\label{sec:results}

\subsection{Single-Client Baselines}

In the single-client setting, each cohort trains independently on its local data,
providing a baseline without federation. Table~\ref{tab:single} reports shared-5
test macro AUC across all four test sets. Performance is dominated by per-dataset
difficulty: PadChest is the easiest test set (most models score highest on it)
and CheXpert the hardest. Only the PadChest-trained model peaks on its own test set; the NIH-trained model
even beats the CheXpert-trained model on CheXpert (0.694 vs.\ 0.636), which is
weakest overall (mean 0.618). We do not attribute this to a single cause. Identical captions
(e.g., ``no findings'') can create contrastive false negatives within a batch, but
a per-batch collision analysis argues against this as the explanation for
CheXpert: of the four cohorts, CheXpert has the \emph{lowest} caption-collision
rate (0.33), whereas NIH and VinDr have the highest ($\geq0.77$) yet yield
stronger models (Table~\ref{tab:single}); if collisions were dominant, this
ordering would be reversed. CheXpert's difficulty
more plausibly reflects its uncertain-label policy and known label-quality
issues~\cite{ref_chexpert,ref_chexpertcomp}.

\begin{table}[h]
\centering
\caption{Single-client baselines: test-set shared-5 macro AUC. Rows are models trained on a single cohort; columns are held-out test sets; diagonal entries in bold.}
\label{tab:single}
\begin{tabular*}{\textwidth}{@{\extracolsep{\fill}}lccccc@{}}
\toprule
Model & NIH & CheXpert & VinDr & PadChest & Mean \\
\midrule
NIH-trained       & \textbf{0.771} & 0.694 & 0.794 & 0.845 & 0.776 \\
CheXpert-trained  & 0.642 & \textbf{0.636} & 0.613 & 0.580 & 0.618 \\
VinDr-trained     & 0.645 & 0.644 & \textbf{0.831} & 0.841 & 0.740 \\
PadChest-trained  & 0.722 & 0.673 & 0.812 & \textbf{0.887} & 0.774 \\
\bottomrule
\end{tabular*}
\end{table}

\subsection{Aggregation Strategy Comparison}

One-shot aggregation merges the four independently trained adapters once, without
further communication rounds, isolating the aggregation method's effect. Naive
FedAvg averaging the $A$ and $B$ factors independently performs poorly (mean
0.684), confirming this algebraically inexact aggregation degrades the merged
model. SVD-based aggregation recovers substantially, reaching 0.781
($+0.097$; Table~\ref{tab:aggregation}). Equal weighting ($w_k=1/N$) outperforms
size weighting ($w_k=n_k/\sum_j n_j$; 0.781 vs.\ 0.743), which lets the largest
cohort dominate. We compare aggregation strategies in the one-shot setting only;
verifying the naive--SVD gap in multi-round federation is left to future work.

\begin{table}[h]
\centering
\caption{One-shot aggregation: test-set shared-5 macro AUC. SVD-based aggregation reconstructs and re-factorizes the update; the size-weighted variant reaches 0.743.}
\label{tab:aggregation}
\begin{tabular*}{\textwidth}{@{\extracolsep{\fill}}lccccc@{}}
\toprule
Aggregation & NIH & CheXpert & VinDr & PadChest & Mean \\
\midrule
Naive FedAvg          & 0.684 & 0.666 & 0.660 & 0.726 & 0.684 \\
SVD (equal-weighted)  & 0.744 & 0.702 & 0.824 & 0.852 & \textbf{0.781} \\
\bottomrule
\end{tabular*}
\end{table}

\subsection{Multi-Round Federation}

In multi-round federation, clients alternate between local training and
SVD-based aggregation for five communication rounds. Table~\ref{tab:federation}
shows the shared-5 test macro AUC after training. FedAvg achieves a mean AUC of
0.802, outperforming both the best single-client baseline (0.776) and one-shot
aggregation (0.781). In our single-seed runs, FedProx performs similarly (0.799),
with overlapping bootstrap CIs across all cohorts, indicating no clear benefit
from the proximal term under LoRA federation. Relative to zero-shot BiomedCLIP (frozen backbone, no adaptation; mean 0.687), federated adaptation improves every cohort $+0.095$ (NIH), $+0.080$ (CheXpert), $+0.170$ (VinDr), $+0.115$ (PadChest) for a mean gain of $+0.115$, confirming that the adaptation, not the pretrained backbone alone, drives performance.

A single model trained on the pooled datasets (3 epochs) reached 0.812.
Federation outperformed it on CheXpert (0.732 vs.\ 0.685) and approached it
overall while keeping data local, though the differing training budgets make the
two not directly comparable.

Full-class evaluation over each cohort's native label set
(Table~\ref{tab:fullclass}) shows the same pattern at lower absolute values
(FedAvg mean 0.734 vs.\ 0.802 shared-5), as the native sets include rarer and
harder findings; FedAvg and FedProx again perform near-identically.

\begin{table}[h]
\centering
\small
\caption{Multi-round federation (5 rounds) vs.\ a centralized reference: test-set shared-5 macro AUC (bootstrap 95\% CIs in brackets). Zero-shot is the frozen BiomedCLIP backbone; the centralized model pools all data under a different schedule and is a reference, not a strict upper bound.}
\label{tab:federation}
\begin{tabular*}{\textwidth}{@{\extracolsep{\fill}}lccccc@{}}
\toprule
Method & NIH & CheXpert & VinDr & PadChest & Mean \\
\midrule
Zero-shot & 0.676 & 0.652 & 0.677 & 0.744 & 0.687 \\
FedAvg  & 0.771{\tiny[0.77,0.78]} & 0.732{\tiny[0.73,0.74]} & 0.847{\tiny[0.81,0.88]} & 0.859{\tiny[0.82,0.89]} & 0.802 \\
FedProx & 0.765{\tiny[0.76,0.77]} & 0.727{\tiny[0.72,0.73]} & 0.845{\tiny[0.81,0.88]} & 0.861{\tiny[0.82,0.89]} & 0.799 \\
Centralized & 0.790 & 0.685 & 0.885 & 0.888 & 0.812 \\
\bottomrule
\end{tabular*}
\end{table}

\begin{table}[h]
\centering
\caption{Full-class evaluation: test-set macro AUC over each cohort's native label set (14 classes for NIH/CheXpert/VinDr, 30 for PadChest) for the multi-round federated models.}
\label{tab:fullclass}
\begin{tabular*}{\textwidth}{@{\extracolsep{\fill}}lccccc@{}}
\toprule
Method & NIH & CheXpert & VinDr & PadChest & Mean \\
\midrule
FedAvg  & 0.722 & 0.713 & 0.754 & 0.747 & 0.734 \\
FedProx & 0.715 & 0.707 & 0.750 & 0.748 & 0.730 \\
\bottomrule
\end{tabular*}
\end{table}

\subsection{Federation versus Single-Client, and Round Dynamics}
\begin{figure}[t]
  \centering
  \includegraphics[width=0.90\textwidth]{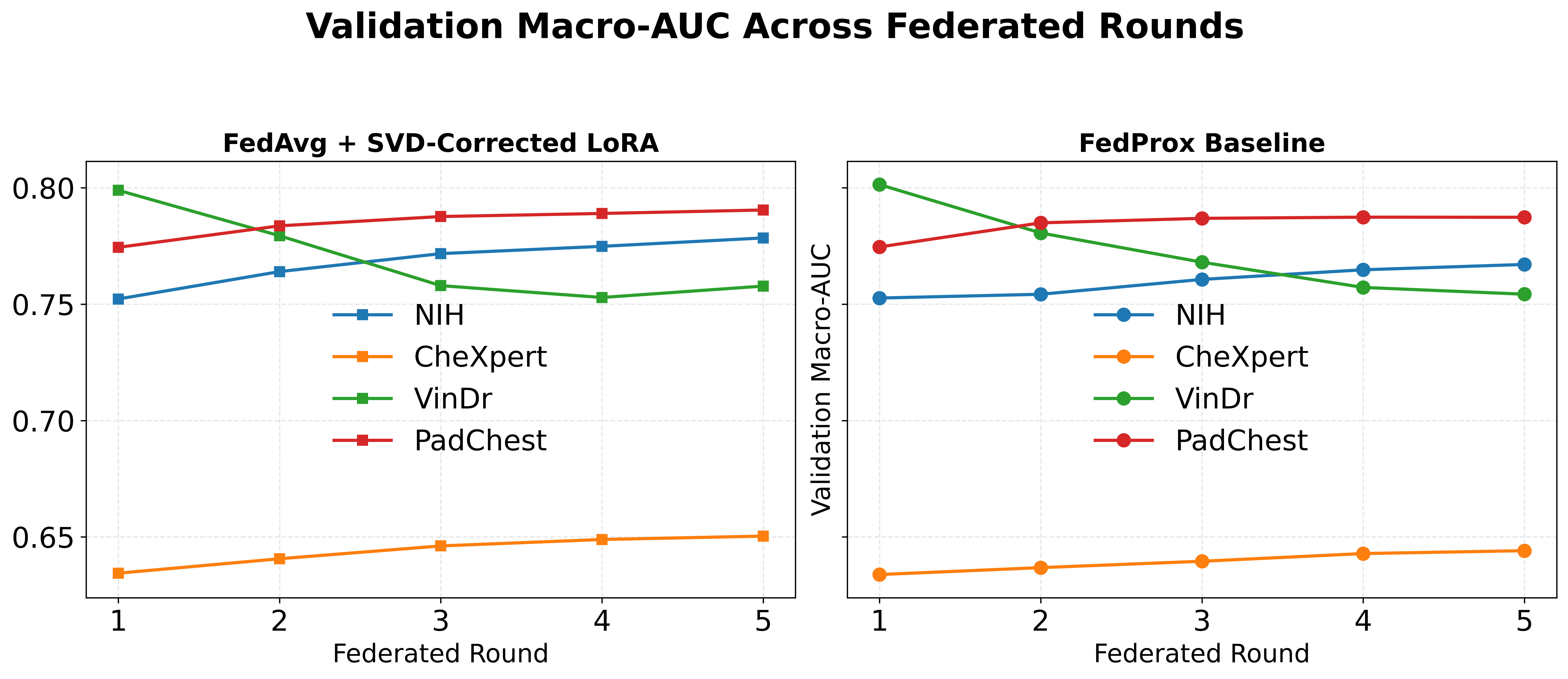}
  \caption{Validation macro AUC (shared-5) per cohort across five rounds: (a) FedAvg, (b) FedProx. The two behave near-identically.}
  \label{fig:rounds}
\end{figure}

Relative to each cohort's strongest single-client baseline, FedAvg improved
CheXpert ($+0.038$) and VinDr ($+0.016$), matched NIH, and reduced PadChest
($-0.028$); the federated mean (0.802) exceeded the best single model's average
(0.776) by $+0.026$. As all results derive from a single seed, small differences (the
VinDr gain, the FedProx--FedAvg gap) should be read as indicative. Validation
curves (Fig.~\ref{fig:rounds}) show stable convergence, with FedAvg and FedProx
nearly identical and VinDr peaking early.

%% ============================================================
%% DISCUSSION
%% ============================================================
\section{Discussion}
\label{sec:discussion}
BiomedCLIP federates effectively across diverse chest radiograph cohorts with parameter-efficient LoRA; two findings stand out.

\noindent\textbf{Federation improves weaker cohorts.}
Gains concentrate on CheXpert and VinDr while the strongest cohort (PadChest) is
largely preserved, extending prior multi-site findings to vision-language
PEFT~\cite{ref_tayebi}.

\noindent\textbf{FedProx provides little benefit.}
In our single-seed runs, FedProx performed similarly to FedAvg. Two factors
plausibly limit client drift: LoRA's low-rank updates constrain the update
subspace, and, with the text encoder frozen, all clients align to identical fixed
prompt embeddings, reducing the divergence a proximal term would correct. As the
penalty acts on the factors rather than their product, we read this as empirical,
not mechanistic~\cite{ref_fedprox,ref_flexlora,ref_fedsa,ref_fedex}.

\noindent\textbf{Limitations.}
Results use a single seed on one accelerator, so CIs reflect test-set not
training variance and comparative gaps (FedProx vs.\ FedAvg) should be read
cautiously. We evaluate five shared findings via macro-AUC without calibration;
FL localizes but does not guarantee privacy; and the image-level VinDr split
cannot exclude within-patient leakage. Future work includes multiple seeds,
per-finding evaluation, more cohorts/backbones, and differential privacy.

%% CONCLUSION
%% ============================================================
\section{Conclusion}
\label{sec:conclusion}

We studied federated PEFT of BiomedCLIP for chest radiograph classification across
four cohorts on three continents. Federated LoRA adaptation improves over the
frozen backbone on all cohorts (mean $0.687\rightarrow0.802$) and approaches a
centralized reference (0.812) without pooling data; SVD-based aggregation is
essential, while FedProx adds no measurable benefit in our single-seed setting.
These results support data-localizing adaptation of biomedical VLMs.


\begin{thebibliography}{99}

  %% ===== Federated learning foundations =====
  \bibitem{ref_fedavg}
  McMahan, B., Moore, E., Ramage, D., Hampson, S., y Arcas, B.A.:
  Communication-efficient learning of deep networks from decentralized data.
  In: AISTATS, pp.\ 1273--1282 (2017)

  \bibitem{ref_fedprox}
  Li, T., Sahu, A.K., Zaheer, M., Sanjabi, M., Talwalkar, A., Smith, V.:
  Federated optimization in heterogeneous networks.
  In: Proceedings of Machine Learning and Systems (MLSys), vol.\ 2, pp.\ 429--450 (2020)

  \bibitem{ref_fedbn}
  Li, X., Jiang, M., Zhang, X., Kamp, M., Dou, Q.:
  FedBN: Federated learning on non-IID features via local batch normalization.
  In: International Conference on Learning Representations (ICLR) (2021)

  \bibitem{ref_scaffold}
  Karimireddy, S.P., Kale, S., Mohri, M., Reddi, S., Stich, S., Suresh, A.T.:
  SCAFFOLD: Stochastic controlled averaging for federated learning.
  In: ICML, pp.\ 5132--5143 (2020)

  \bibitem{ref_fedsurvey}
  Kairouz, P., McMahan, H.B., Avent, B., Bellet, A., Bennis, M., et al.:
  Advances and open problems in federated learning.
  Foundations and Trends in Machine Learning \textbf{14}(1--2), 1--210 (2021)

  %% ===== LoRA and federated LoRA =====
  \bibitem{ref_lora}
  Hu, E.J., Shen, Y., Wallis, P., Allen-Zhu, Z., Li, Y., Wang, S., Wang, L., Chen, W.:
  LoRA: Low-rank adaptation of large language models.
  In: International Conference on Learning Representations (ICLR) (2022)

  \bibitem{ref_fedit}
  Zhang, J., Vahidian, S., Kuo, M., Li, C., Zhang, R., Yu, T., Wang, G., Chen, Y.:
  Towards building the federated GPT: Federated instruction tuning.
  In: ICASSP (2024)

  \bibitem{ref_flexlora}
  Bai, J., Chen, D., Qian, B., Yao, L., Li, Y.:
  Federated fine-tuning of large language models under heterogeneous tasks and client resources.
  In: Advances in Neural Information Processing Systems (NeurIPS), vol.\ 37 (2024)

  \bibitem{ref_fedsa}
  Guo, P., Zeng, S., Wang, Y., Fan, H., Wang, F., Qu, L.:
  Selective aggregation for low-rank adaptation in federated learning.
  In: ICLR (2025)

  \bibitem{ref_fedex}
  Singhal, R., Ponkshe, K., Vepakomma, P.:
  FedEx-LoRA: Exact aggregation for federated and efficient fine-tuning of foundation models.
  In: Annual Meeting of the Association for Computational Linguistics (ACL), pp.\ 1316--1336 (2025)

  \bibitem{ref_ffalora}
  Sun, Y., Li, Z., Li, Y., Ding, B.:
  Improving LoRA in privacy-preserving federated learning.
  In: International Conference on Learning Representations (ICLR) (2024)

  \bibitem{ref_poudel2026}
  Poudel, S., Kunwor, N., Simkhada, R., Munir, M., Dhakal, M., Poudel, K.:
  Parameter-Efficient Fine-Tuning for Domain-Specific Gastrointestinal Disease Recognition.
  In: CVPRW (2026)

  %% ===== Vision-language models =====
  \bibitem{ref_clip}
  Radford, A., Kim, J.W., Hallacy, C., Ramesh, A., Goh, G., Agarwal, S., Sastry, G., Askell, A., Mishkin, P., Clark, J., Krueger, G., Sutskever, I.:
  Learning transferable visual models from natural language supervision.
  In: International Conference on Machine Learning (ICML), pp.\ 8748--8763 (2021)

  \bibitem{ref_biomedclip}
  Zhang, S., Xu, Y., Usuyama, N., Xu, H., Bagga, J., Tinn, R., Preston, S., Rao, R., Wei, M., Valluri, N., Wong, C., Tupini, A., Wang, Y., Mazzola, M., Shukla, M., Liden, L., Gao, J., Lungren, M.P., Naumann, T., Wang, S., Poon, H.:
  BiomedCLIP: A multimodal biomedical foundation model pretrained from fifteen million scientific image-text pairs.
  arXiv:2303.00915 (2023)

  \bibitem{ref_medclip}
  Wang, Z., Wu, Z., Agarwal, D., Sun, J.:
  MedCLIP: Contrastive learning from unpaired medical images and text.
  In: Conference on Empirical Methods in Natural Language Processing (EMNLP), pp.\ 3876--3887 (2022)

  \bibitem{ref_chexzero}
  Tiu, E., Talius, E., Patel, P., Langlotz, C.P., Ng, A.Y., Rajpurkar, P.:
  Expert-level detection of pathologies from unannotated chest X-ray images via self-supervised learning.
  Nature Biomedical Engineering \textbf{6}, 1399--1406 (2022)

  \bibitem{ref_biovil}
  Boecking, B., Usuyama, N., Bannur, S., Castro, D.C., Schwaighofer, A., Hyland, S., Wetscherek, M., Naumann, T., Nori, A., Alvarez-Valle, J., Poon, H., Oktay, O.:
  Making the most of text semantics to improve biomedical vision-language processing.
  In: European Conference on Computer Vision (ECCV), pp.\ 1--21 (2022)

  \bibitem{ref_vlsmadapter}
  Dhakal, M., Adhikari, R., Thapaliya, S., Khanal, B.:
  VLSM-Adapter: Finetuning vision-language segmentation efficiently with lightweight blocks.
  In: Medical Image Computing and Computer-Assisted Intervention (MICCAI), pp. 1--13 (2024)

  %% ===== Federated CLIP / foundation models =====
  \bibitem{ref_fedclip}
  Lu, W., Hu, X., Wang, J., Xie, X.:
  FedCLIP: Fast generalization and personalization for CLIP in federated learning.
  IEEE Data Engineering Bulletin \textbf{46}(1), 52--66 (2023)

  \bibitem{ref_fedfms}
  Liu, Y., Luo, G., Zhu, Y., Feng, Q., Chen, T., Liu, Q.:
  FedFMS: Exploring federated foundation models for medical image segmentation.
  In: Medical Image Computing and Computer-Assisted Intervention (MICCAI), pp.\ 283--293 (2024)

  %% ===== Chest radiograph datasets =====
  \bibitem{ref_nih}
  Wang, X., Peng, Y., Lu, L., Lu, Z., Bagheri, M., Summers, R.M.:
  ChestX-ray8: Hospital-scale chest X-ray database and benchmarks on weakly-supervised classification and localization of common thorax diseases.
  In: IEEE Conference on Computer Vision and Pattern Recognition (CVPR), pp.\ 2097--2106 (2017)

  \bibitem{ref_chexpert}
  Irvin, J., Rajpurkar, P., Ko, M., Yu, Y., Ciurea-Ilcus, S., Chute, C., Marklund, H., Haghgoo, B., Ball, R., Shpanskaya, K., et al.:
  CheXpert: A large chest radiograph dataset with uncertainty labels and expert comparison.
  In: AAAI Conference on Artificial Intelligence, pp.\ 590--597 (2019)

  \bibitem{ref_vindr}
  Nguyen, H.Q., Lam, K., Le, L.T., Pham, H.H., Tran, D.Q., Nguyen, D.B., Le, D.D., Pham, C.M., Tong, H.T.T., Dinh, D.H., et al.:
  VinDr-CXR: An open dataset of chest X-rays with radiologist's annotations.
  Scientific Data \textbf{9}, 429 (2022)

  \bibitem{ref_padchest}
  Bustos, A., Pertusa, A., Salinas, J.M., de la Iglesia-Vay\'a, M.:
  PadChest: A large chest x-ray image dataset with multi-label annotated reports.
  Medical Image Analysis \textbf{66}, 101797 (2020)

  \bibitem{ref_mimic}
  Johnson, A.E.W., Pollard, T.J., Berkowitz, S.J., Greenbaum, N.R., Lungren, M.P., Deng, C., Mark, R.G., Horng, S.:
  MIMIC-CXR, a de-identified publicly available database of chest radiographs with free-text reports.
  Scientific Data \textbf{6}, 317 (2019)

  %% ===== Federated learning for medical imaging =====
  \bibitem{ref_tayebi}
  Tayebi Arasteh, S., Kuhl, C., Saehn, M.J., Isfort, P., Truhn, D., Nebelung, S.:
  Enhancing domain generalization in the AI-based analysis of chest radiographs with federated learning.
  Scientific Reports \textbf{13}, 22576 (2023)

  \bibitem{ref_surgical}
  Kulkarni, P., Kanhere, A., Yi, P.H., Parekh, V.S.:
  From isolation to collaboration: Federated class-heterogeneous learning for chest X-ray classification.
  arXiv:2301.06683 (2023)

  \bibitem{ref_fedmed}
  Yan, Z., Wicaksana, J., Wang, Z., Yang, X., Cheng, K.T.:
  Variation-aware federated learning with multi-source decentralized medical image data.
  IEEE Journal of Biomedical and Health Informatics \textbf{25}(7), 2615--2628 (2021)

  \bibitem{ref_chexpertcomp}
  Cohen, J.P., Hashir, M., Brooks, R., Bertrand, H.:
  On the limits of cross-domain generalization in automated X-ray prediction.
  In: Medical Imaging with Deep Learning (MIDL), pp.\ 136--155 (2020)

  \bibitem{ref_dp}
  Ziller, A., Usynin, D., Braren, R., Makowski, M., Rueckert, D., Kaissis, G.:
  Medical imaging deep learning with differential privacy.
  Scientific Reports \textbf{11}, 13524 (2021)

  \bibitem{ref_fedsvd}
  Lee, S., Park, S., Lee, D.B., Wagner, D., Seong, H., Bocklet, T., Lee, J., Hwang, S.J.:
  FedSVD: Adaptive orthogonalization for private federated learning with LoRA.
  arXiv:2505.12805 (2025)

\end{thebibliography}
\end{document}